\documentclass{article}

\usepackage{rfPRIMEarxiv}
\usepackage[numbers,sort&compress]{natbib}
\usepackage[utf8]{inputenc} % allow utf-8 input
\usepackage[T1]{fontenc}    % use 8-bit T1 fonts
\usepackage{url}            % simple URL typesetting
\usepackage{booktabs}       % professional-quality tables
\usepackage{amsfonts}       % blackboard math symbols
\usepackage{nicefrac}       % compact symbols for 1/2, etc.
\usepackage{microtype}      % microtypography
\usepackage{lipsum}
\usepackage{threeparttable}
\usepackage{graphicx}
\usepackage{tabularx}
\usepackage{multirow}
\usepackage{fancyhdr}       % header
\usepackage{graphicx}       % graphics
\graphicspath{{media/}}     % organize your images and other figures under media/ folder
\usepackage{booktabs}
\usepackage{tabularx}
\usepackage{caption}
\usepackage{array}
\usepackage{xcolor}
\usepackage{soul}
\usepackage{amsmath}      % For \textbf and other math functions
\usepackage{textcomp}     % For \texttildelow
\usepackage{hyperref}       % hyperlinks
\sethlcolor{yellow}
\title{From Embeddings to Accuracy: Comparing Foundation Models for Radiographic Classification
}

\author{%
  \parbox[t]{\textwidth}{\centering
    Xue Li\textsuperscript{1}, Jameson Merkow\textsuperscript{2}, Noel C. F. Codella\textsuperscript{2}, Alberto Santamaria-Pang\textsuperscript{2}, \\
    \textbf{Naiteek Sangani\textsuperscript{2}, Alexander Ersoy\textsuperscript{2}, Christopher Burt\textsuperscript{2}, John W. Garrett\textsuperscript{1}, Richard J. Bruce\textsuperscript{1}}, \\
    \textbf{Joshua D. Warner\textsuperscript{1}, Tyler Bradshaw\textsuperscript{1}, Ivan Tarapov\textsuperscript{2}, Matthew P. Lungren\textsuperscript{2}, Alan B. McMillan\textsuperscript{1}}\\[3ex]
    \normalfont
    \textsuperscript{1}University of Wisconsin–Madison, Madison, WI USA\\
    \textsuperscript{2}Microsoft Health and Life Sciences, Redmond, WA USA
  }
}

\begin{document}
\maketitle

\begin{abstract}
Foundation models, pretrained on extensive datasets, have significantly advanced machine learning by providing robust and transferable embeddings applicable to various domains, including medical imaging diagnostics. This study evaluates the utility of embeddings derived from both general-purpose and medical domain-specific foundation models for training lightweight adapter models in multi-class radiography classification, focusing specifically on tube placement assessment and related findings. A dataset comprising 8,842 radiographs classified into seven distinct categories was employed to extract embeddings using seven foundation models: DenseNet121, BiomedCLIP, Med-Flamingo, MedImageInsight, MedSigLIP, Rad-DINO, and CXR-Foundation. Adapter models were subsequently trained using classical machine learning algorithms, including K-Nearest Neighbors (KNN), logistic regression (LR), Support Vector Machines (SVM), random forest (RF), and Multi-Layer Perceptron (MLP). Among these combinations, MedImageInsight embeddings paired with an SVM or MLP adapter yielded the highest mean area under the curve (mAUC) at 93.1\%, followed closely by MedSigLIP with MLP (91.0\%), Rad-DINO with SVM (90.7\%), and CXR-Foundation with LR (88.6\%). In comparison, BiomedCLIP and DenseNet121 exhibited moderate performance with SVM, obtaining mAUC scores of 82.8\% and 81.1\%, respectively, whereas Med-Flamingo delivered the lowest performance at 78.5\% when combined with RF. Significant differences were found between each embedding model and MedImageInsight using the Wilcoxon signed-rank test at the significance level 0.05 (before Bonferroni correction). Notably, most adapter models demonstrated computational efficiency, achieving training within minutes and inference within seconds on CPU, underscoring their practicality for clinical applications. Furthermore, fairness analysis on adapters trained on MedImageInsight-derived embeddings indicated minimal disparities, with gender differences in performance within 1.8\% and standard deviations across age groups not exceeding 1.4\%. Further analysis indicated there is no significant difference across gender and age at significance level 0.05. These findings confirm that foundation model embeddings—especially those from MedImageInsight—facilitate accurate, computationally efficient, and equitable diagnostic classification using lightweight adapters for radiographic image analysis.
\end{abstract}

% keywords can be removed

\section{Introduction}
Radiography remains an essential diagnostic imaging modality, playing a critical role in clinical decision-making, procedural guidance, and patient monitoring \cite{wu2023towards}. Accurate interpretation of radiographs, especially for verifying correct tube placement and findings, is directly tied to patient safety and treatment outcomes, as misplacement can lead to severe complications \cite{keyte2021immediate, mcmullen2022nasogastric}. Despite its importance, radiology faces an increasing burden due to the immense volume of imaging studies. In the United States alone, approximately 5,000 radiologists are tasked with interpreting over 126 million radiographic examinations each year \cite{IMV2021}. This heavy workload can strain radiologists, potentially impacting diagnostic accuracy, efficiency, and overall patient care quality.

In recent years, artificial intelligence (AI) and machine learning (ML) have been progressively integrated into radiological practice to enhance diagnostic accuracy and workflow efficiency \cite{Kelly2022,Mello-Thoms2023,Najjar2023}. Despite these promising developments, the implementation of end-to-end deep learning models in clinical practice often faces significant challenges. They inherently require task-specific training \cite{kniggemodelling}. Consequently, whenever a new clinical condition or classification task arises, developing and training an entirely new model becomes necessary. Additionally, these approaches face significant challenges in generalizing across different sites due to variations in imaging protocols and equipment, which shift data distributions \cite{ahmed2021discovery, salehi2023study}. As a result, each medical site typically needs to independently train new models, an approach that is both resource-intensive and inefficient. These limitations underscore the urgent need for more flexible and resource-efficient solutions.

More recently, foundation models represent a significant advancement in artificial intelligence, characterized by large-scale architectures pre-trained through self-supervised learning on extensive, multimodal datasets \cite{moor2023foundation, zhou2024comprehensive, awais2025foundation}. A primary strength of foundation models is their ability to produce generalizable embeddings which capture a compact, high-dimensional representation that is rich with meaningful information from input data. These embeddings can be efficiently adapted or fine-tuned for a wide range of downstream tasks, eliminating the need to train entirely new models from scratch \cite{radford2021learning, zhang2024biomedclip, Paschali2025}. Consequently, integrating foundation model embeddings with traditional machine learning classifiers presents a novel and promising approach in radiographic image analysis \cite{Willemink2022}. Utilizing these robust feature representations, conventional machine learning classifiers such as k-nearest neighbors (KNN), logistic regression (LR), support vector machines (SVM), random forests (RF), and multi-layer perceptrons (MLP) can be effectively optimized for multi-class classification. Preliminary results indicate that this integration can achieve high mean area under the curve (mAUC) metrics \cite{ zhang2024biomedclip, Sellergren2022, codella2024medimageinsight}, demonstrating strong potential in detecting subtle radiographic variations essential for accurate tube placement verification and findings identification. Nevertheless, a systematic evaluation comparing embeddings derived from general-purpose versus medical domain-specific models is critical to determine the most effective embedding strategy.

This study seeks to address this gap by systematically assessing the performance of various foundation model embeddings in training lightweight adapter classifiers for multi-class classification of radiography images. By focusing on tube placement and related findings, a task with significant clinical implications, we aim to determine the optimal combinations that not only enhance diagnostic accuracy but also ensure computational efficiency. The evaluation spans multiple models and classifiers, providing a comprehensive comparison that reflects the diverse capabilities of current foundation models in a clinically relevant setting. We hypothesize that these embeddings will enable high diagnostic performance while maintaining the computational efficiency necessary for real-time clinical deployment.

\section{Materials and Methods}

\subsection{Study Design and Dataset}

This study was designed to evaluate the performance of lightweight adapter models trained on embeddings derived from pre-trained foundation models for the multi-class classification of radiography images, with a focus on tube placement verification and findings identification. The dataset was developed to support tasks where rapid and automated feedback is clinically valuable, such as the evaluation of tube and line presence/placement, pneumothorax detection, and the inclusion of normal controls. Cases were initially identified through keyword searches of the radiology report text and study order indications. Final class labels were then assigned based on joint review of the reports by a medical physicist and a board-certified radiologist, with the adjudicated report serving as the reference standard. 

\begin{figure}[t]
    \centering
    \includegraphics[width=0.92\textwidth]{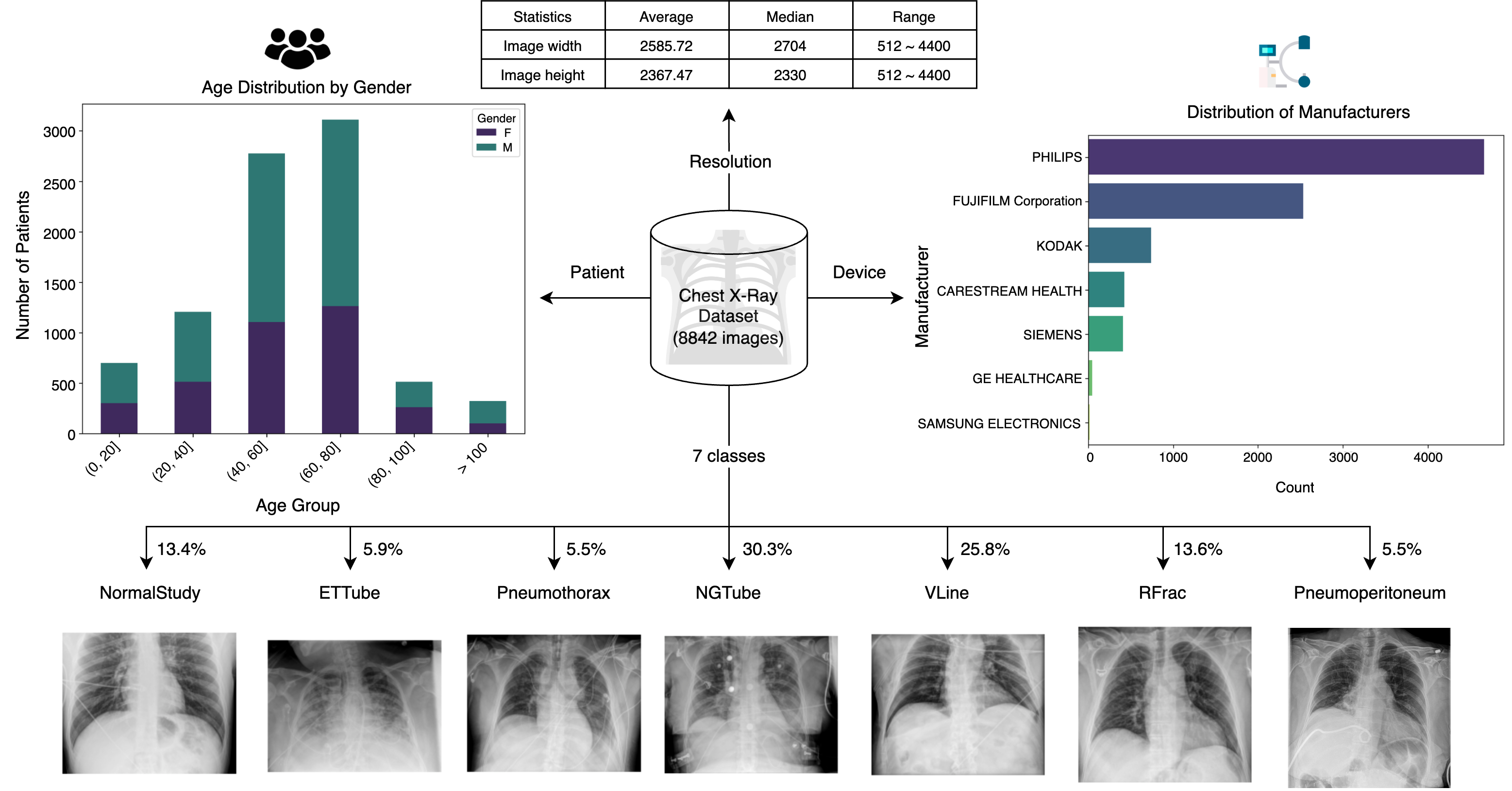}
    \caption{Overview of the Chest X-ray dataset used in this study. The dataset comprises 8,842 radiographs, categorized into seven diagnostic classes: NormalStudy (13.4\%), ETTube (5.9\%), Pneumothorax (5.5\%), NGTube (30.3\%), VLine (25.8\%), RFrac (13.6\%), and Pneumoperitoneum (5.5\%). (\textbf{Left}) The age distribution of patients, stratified by gender. (\textbf{Right}) The distribution of imaging device manufacturers. (\textbf{Top}) Summary of image resolution statistics.}
    \label{fig:dataset}
\end{figure}

As summarized in Fig. \ref{fig:dataset}, this dataset comprises 8,842 radiographs from 7045 patients categorized into seven diagnostic categories, including endotracheal tube (ETTube, 5.9\%) and nasogastric tube (NGTube, 30.3\%) placements, a normal radiographic study (NormalStudy, 13.4\%), and three critical pathological findings—air in the peritoneal cavity (pneumoperitoneum, 5.5\%), air in the pleural space (pneumothorax, 5.5\%), and rib fractures (RFrac, 13.6\%)—along with the presence of a vascular line (VLine, 25.8\%), such as a central venous catheter or peripherally inserted central catheter (PICC line). These classes are relatively uncommon in public datasets, making this dataset well-suited for evaluating the generalization capabilities of foundation model embeddings.

The cases were retrospectively collected, encompassing a diverse range of patient demographics and imaging device manufacturers, as illustrated in Fig. \ref{fig:dataset}. The dataset includes radiographs from patients across various age groups, with a higher proportion in the 40–80 age range, and imaging data acquired from multiple manufacturers, including Philips, Fujifilm, and Kodak, among others. All cases originated from computed radiography (CR) or digital radiography (DR) systems. The references to Fujifilm and Kodak reflect the CR cassettes used during image acquisition rather than digitized film; no film-based images were included. Accordingly, the dataset reflects contemporary digital practice rather than legacy film workflows. This diversity ensures a broad representation of clinical scenarios relevant to tube placement assessment and findings. Due to the retrospective nature of this study, a waiver of informed consent was obtained.

\subsection{Image Preprocessing and Embedding Extraction}

\begin{figure}[h!]
    \centering
    \includegraphics[width=0.92\textwidth]{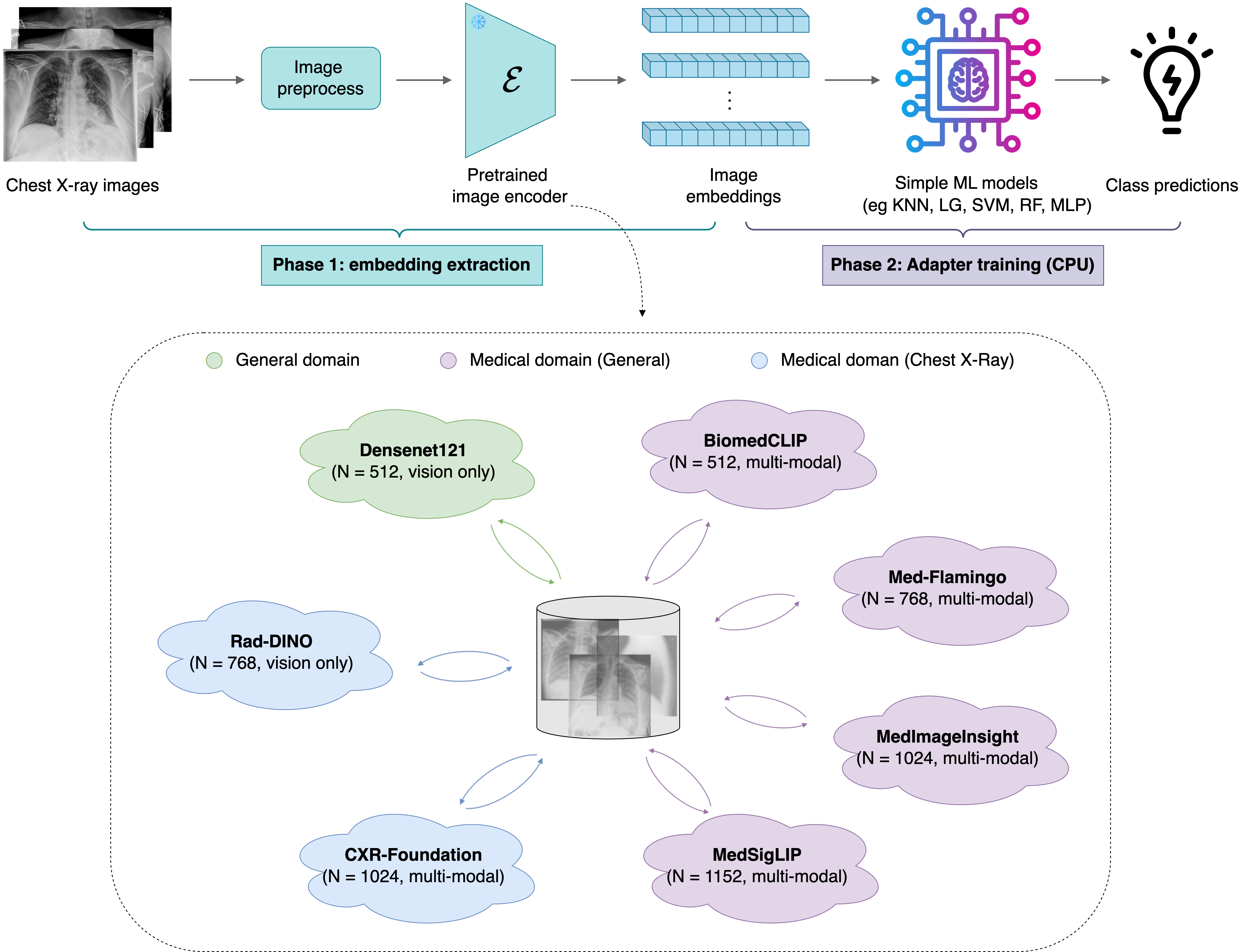}
    \caption{Two-phase workflow for radiograph classification. Phase 1: Embedding Extraction—Chest X-ray images were preprocessed and passed through pre-trained foundation models, including DenseNet121, BiomedCLIP, Med-Flamingo, MedImageInsight, MedSigLIP, Rad-DINO, and CXR-Foundation, to generate image embeddings. Phase 2: Adapter Training—Embeddings were used as input to lightweight ML models for class prediction, enabling efficient transfer learning.}
    \label{fig:workflow}
\end{figure}

Each radiograph underwent a standardized preprocessing protocol to ensure uniformity across the dataset. Initially, the images were normalized by rescaling pixel intensities to a fixed range of 0 to 255. This normalization step was critical to minimize variations due to differing acquisition parameters and to facilitate a consistent input format for subsequent processing. The normalized images were then passed through seven distinct pre-trained foundation models— DenseNet-121 \cite{huang2017densely}, BiomedCLIP \cite{zhang2024biomedclip}, Med-Flamingo \cite{moor2023med}, MedImageInsight \cite{codella2024medimageinsight}, MedSigLIP \cite{sellergren2025medgemma}, Rad-DINO \cite{Perez-Garcia2025}, and CXR-Foundation\cite{Sellergren2022}—to extract high-dimensional feature embeddings as shown in Phase 1 of Fig. \ref{fig:workflow}. These models, pre-trained on large-scale datasets, functioned as automated feature extractors, generating representations that capture the intricate structural and textural details of the radiographs. DenseNet was sourced from \textit{timm} \cite{rw2019timm} and the remaining models from HuggingFace \cite{Microsoft_BiomedCLIP,Google_CXRFoundation,TeamMF_MedFlamingo,Microsoft_MedImageInsight,Google_MedSigLIP,Microsoft_RadDINO}. Embedding extraction was performed using the default configurations provided by each model’s developers to maintain consistency and reproducibility across the embedding extraction pipeline. The embedding length is 512 for BiomedCLIP; 768 for Med-Flamingo and Rad-DINO; 1024 DenseNet-121, MedImageInsight, and CXR-Foundation; and 1152 for MedSigLIP. To analyze the learned representations from each foundation model before adapter training, t-SNE \cite{van2008visualizing} was applied to the extracted embeddings for visualization.

\subsection{Adapter Model Training and Optimization}

The resulting feature embeddings served as inputs for training a series of lightweight adapter classifiers as illustrated in Phase 2 of Fig. \ref{fig:workflow}. The dataset was partitioned into three non-overlapping subsets: a training set, a validation set, and an independent test set with ratios of 64\%, 16\%, and 20\%, respectively. The training set was employed to fit various traditional machine learning classifiers using \textit{scikit-learn}\cite{Pedregosa2011}: K-nearest neighbors (KNN), logistic regression (LG), support vector machines (SVM), random forest (RF), and multi-layer perceptron (MLP). For each classifier, hyperparameter optimization was carried out using the validation set. This optimization process involved systematic adjustments to key parameters—such as the number of neighbors in the KNN algorithm, optimization methods in the LG algorithm, kernel functions and its degree in SVM, tree depth and number of trees in the RF, and the architectural parameters in the MLP—to maximize performance. Once the optimal hyperparameters were determined, the final models were retrained on the training data with the optimal hyperparameters before being evaluated on the test set. For statistical significance analysis, five-fold cross-validation was used with optimal parameters. In each fold, one subset was held out as the test set, and the remaining four were used for training and validation. This process was repeated five times, with performance metrics averaged across all folds.

\subsection{Performance Evaluation}

Model performance was primarily evaluated using the mAUC, calculated for each combination of foundation model embeddings and lightweight adapter models. For a comprehensive assessment, additional metrics—including accuracy, precision, recall, and F1 score—were computed for each diagnostic class using the best-performing embedding-adapter combination, defined by the highest overall mAUC. To facilitate deeper insights into classification performance, confusion matrices were generated to visualize class-specific predictions and to identify frequent misclassification patterns.

Beyond overall accuracy, fairness was systematically analyzed across gender and age groups for the optimal embedding model combined with various adapter configurations. Specifically, area under the curve (AUC) scores were calculated separately for females (2952 patients) and males (4090 patients), as well as across five distinct age brackets: (0, 20], (20, 40], (40, 60], (60, 80], and (80, 100] with 640, 909, 2161, 2525, and 418 patients respectively. Patients older than 100 were unidentified and were excluded from analysis.

Additionally, statistical analyses were performed to assess whether significant differences existed across embedding models, gender, and age groups. For comparisons between embedding models, the Wilcoxon signed-rank test was applied to evaluate differences between each embedding model and the one performing best, with a significance threshold of 0.05 (adjusted to 0.008 using Bonferroni correction). For gender, which included only two groups, the Wilcoxon signed-rank test was likewise employed, with a significance level of 0.05 (adjusted to 0.01 with Bonferroni correction). In contrast, for the five predefined age groups, the Kruskal–Wallis H-test was used to detect differences, with a significance level of 0.05 (adjusted to 0.01 with Bonferroni correction).

Finally, computational efficiency was evaluated by measuring training and inference times (in seconds) for each adapter model executed on a CPU. These efficiency measurements underscore the lightweight nature of the adapters, highlighting their suitability for rapid deployment, particularly in resource-constrained environments. This evaluation framework rigorously validates the hypothesis that foundation model embeddings can effectively train lightweight adapters for multi-class radiograph classification, ensuring high diagnostic accuracy, computational efficiency, and equitable performance across demographic groups, thus supporting reliable and fair clinical practice.

\section{Results}
\begin{figure}[h!]
    \centering
    \includegraphics[width=0.92\textwidth]{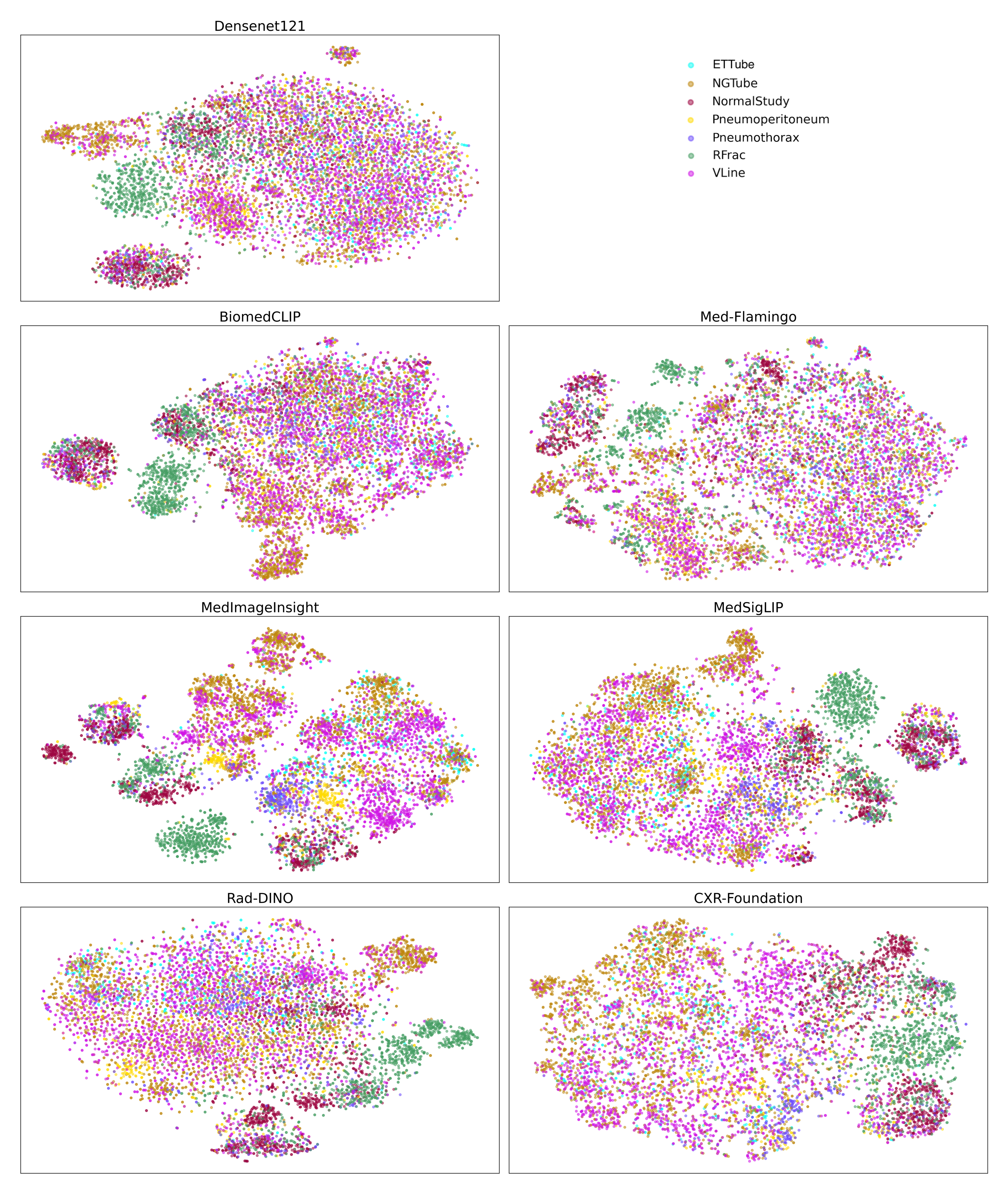}
    \caption{t-SNE visualization of embeddings extracted from seven foundation models: DenseNet121, BiomedCLIP, Med-Flamingo, MedImageInsight, MedSigLIP, Rad-DINO, and CXR-Foundation. Each point represents a radiograph, color-coded according to its diagnostic category: ETTube, NGTube, NormalStudy, Pneumoperitoneum, Pneumothorax, RFrac, and VLine.}
    \label{fig:visualization}
\end{figure}

A total of 8,842 radiographs were analyzed, each labeled into one of seven diagnostic categories relevant to tube placement verification and findings. These images were first processed and then fed into the pre-trained foundation models. The visualization of the embeddings from seven foundation models involved in this study is shown in Fig. \ref{fig:visualization}. The clustering patterns highlighted each model’s ability to differentiate diagnostic categories within the embedding space. Notably, MedImageInsight embeddings exhibited distinct separation across several clinical categories, indicating their effectiveness in capturing subtle yet critical visual differences essential for accurate radiographic classification. Similarly, CXR-Foundation embeddings demonstrated promising clustering patterns, with observable transitions among classes. In contrast, embeddings from other models lacked clear separation, suggesting differences in how each model encodes radiographic features. 

\begin{figure}[t!]
    \centering
    \includegraphics[width=0.92\textwidth]{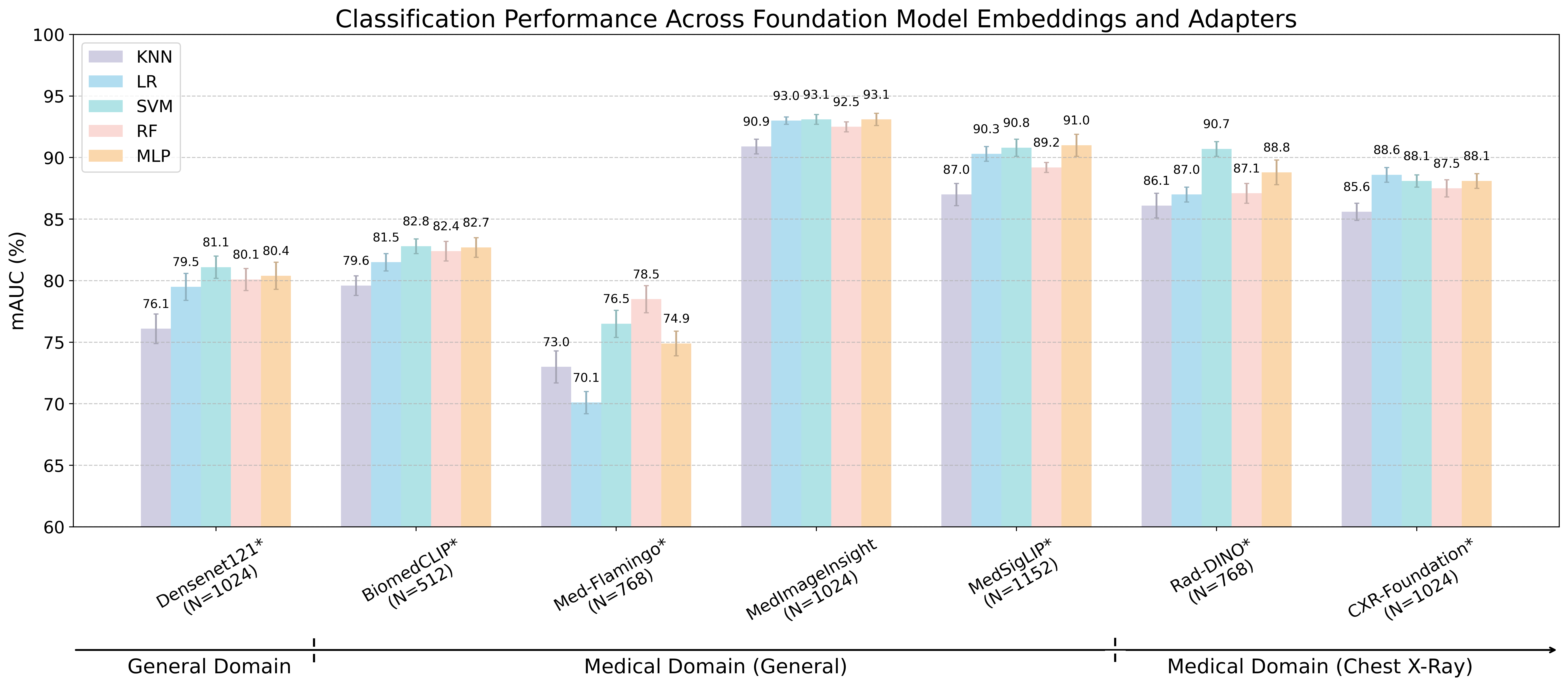}
    \caption{Performance comparison of adapter models (KNN, LR, SVM, RF, and MLP) paired with embeddings from general-purpose to domain-specific foundation models, evaluated by mAUC where error bars reflect standard deviation. \textit{N} denotes the embedding size used for adapter training.}
    \label{fig:results}
\end{figure}

Once the embeddings were extracted, lightweight adapters were trained on CPU for radiographic classification. The classification performance of all combinations of foundation model embeddings and lightweight adapter models was evaluated using mAUC, as shown in Fig. \ref{fig:results}. Across configurations, the adapters exhibited a range of diagnostic performance, with notable variation driven by the choice of foundation model embeddings. Among the seven foundation models, MedImageInsight consistently achieved the highest overall mAUC across adapter types, with top performance from SVM and MLP (93.1\%), followed closely by LR (93.0\%) and RF (92.5\%). Additionally, it also attained the lowest overall uncertainty calculated from standard deviation of each fold for each adapter, 0.6\% for KNN, 0.3\% for LR, 0.4\% for SVM and RF, and 0.5\% for MLP. In contrast, Med-Flamingo produced the lowest mAUC scores, with all adapters falling below 79\%, and higher uncertainty, ranging from 0.9\% to 1.3\%, reflecting its limited suitability for this task and highlighting the variability in embedding quality across foundation models.

When evaluating foundation models, a clear trend emerges along the spectrum from general to domain-specific. DenseNet121, originally trained on broad natural image datasets, demonstrated moderate performance (mAUC range: 76.1\%–81.1\%, uncertainty range: 0.9\%-1.2\%). Models with some biomedical adaptation, like BiomedCLIP (trained on scientific figure–caption pairs, mAUC 79.6\%–82.7\%) and uncertainty (0.6\%-0.8\%), showed incremental improvements but remained limited for this specific radiographic task. Surprisingly, Med-Flamingo, another medical domain–adapted model trained on scientific image–text pairs, achieved an mAUC of only 70.1\%–78.5\% and uncertain of 0.9\%-1.3\%, which is worse than DenseNet121.  In contrast, models more closely aligned with radiographic imaging outperformed these approaches: Rad-DINO (focused on radiography images, mAUC 86.1\%–90.7\%, and uncertainty 0.6\%-1.0\%) and CXR-Foundation (specialized in chest X-rays, mAUC 85.6\%–88.6\%, and uncertainty 0.5\%-0.7\%). Notably, MedImageInsight and MedSigLIP—though not tailored to any single task—achieved the relatively higher mAUC and lower uncertainty, (90.0\%-93.1\%, 0.3\%-0.6\%) and (87.0\%-91.0\%, 0.4\%-0.9\%) respectively, likely owing to their broad domain-specific training across multiple medical imaging modalities. Significant differences were observed between MedImageInsight and each of other embedding models based on the p-values calculated from the Wilcoxon signed-rank test.

From an adapter model perspective, SVM consistently ranked first or second among the seven foundation models, underscoring their strong compatibility with high-dimensional embeddings. LR, RF, and MLP also delivered competitive results, frequently trailing SVM by only a small margin. In contrast, KNN generally exhibited lower performance, particularly when paired with weaker embeddings. Overall, the performance gap among adapter models was much smaller than the disparity observed among embeddings.

\begin{figure}[t!]
    \centering
    \includegraphics[width=0.92\textwidth]{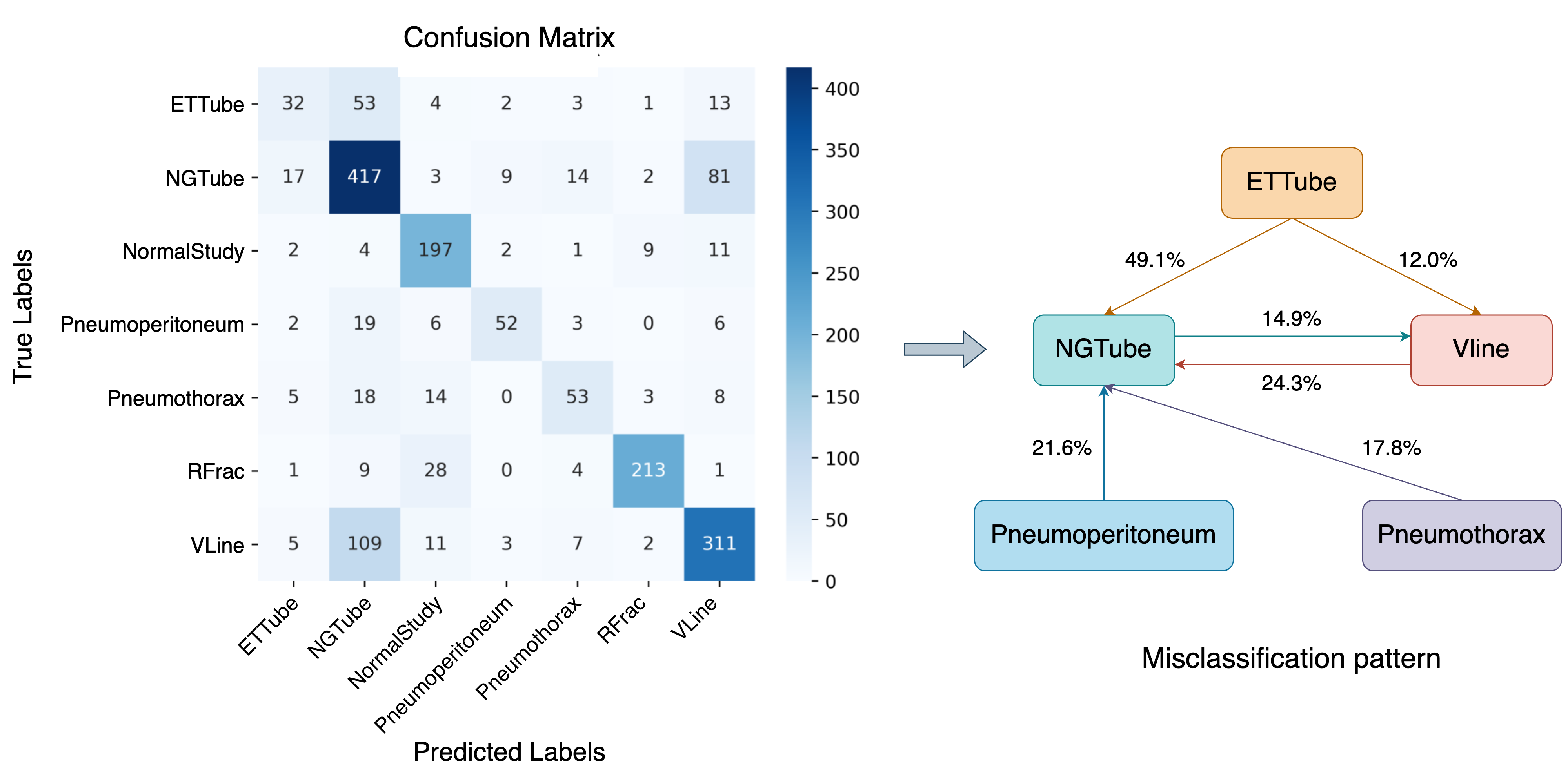}
    \caption{Confusion matrix (left) and corresponding misclassification pattern diagram (right) for multi-class radiographic image classification using MedImageInsight embeddings paired with an SVM adapter. The confusion matrix illustrates counts of predictions versus true labels, highlighting correctly classified samples along the diagonal and misclassifications off-diagonal. The misclassification pattern diagram indicates prevalent confusion among specific classes, with arrow directions representing the direction of misclassification and percentages denoting the proportion of misclassified cases relative to the total cases of each true class.}
    \label{fig:confusion_matrix}
\end{figure}

The combination of MedImageInsight embeddings with an SVM adapter achieved the best overall performance. Detailed
per-class results for this combination are presented in Table \ref{tab:deep_dive_table}, with mAUC values exceeding 90\% in six of the seven diagnostic categories. RFrac achieved the highest mAUC (98.1\%) and F1 score (87.7\%), while ETTube proved
the most challenging, with a low recall (29.6\%) and F1 score (37.2\%), reflecting frequent misclassifications. The confusion matrix in Fig. \ref{fig:confusion_matrix} further reveals several notable misclassification patterns. Misclassification of ETTube as NGTube represents the largest single source of error (49.1\%), with a further 12.0\%) of ETTube cases confused with VLine. Substantial bidirectional confusion also arises between NGTube and VLine (14.9\% versus 24.3\%, respectively). Moreover, 21.6\% of Pneumoperitoneum and 17.8\% of Pneumothorax instances are incorrectly classified as NGTube, indicating that the model occasionally interprets radiographic features of pathological air collections as tube structures.
Overall, these patterns highlight the difficulty in differentiating visually similar lines and tubes (e.g., ETTube, NGTube, and VLine) on chest radiographs, suggesting that finer-grained radiographic features are crucial for accurate classification.

\begin{table}[t]
    \centering
    \caption{Per-class performance (\%) of the SVM adapter with MedImageInsight embeddings}
    \vspace{2mm}
    \renewcommand{\arraystretch}{1.2} % Adjust row height for better readability
    \newcolumntype{C}[1]{>{\centering\arraybackslash}p{#1}} % Define centered column type
    \begin{threeparttable}
    \begin{tabular}{l C{3.2cm} C{2.8cm} C{2.8cm} C{2.8cm}}  
        \toprule
        \textbf{Class} & \textbf{Precision} & \textbf{Recall} & \textbf{F1} & \textbf{AUC} \\
        \midrule
        ETTube          & 50.0  & 29.6  & 37.2  & 90.3  \\
        NGTube          & 66.3  & 76.8  & 71.2  & 89.1  \\
        Pneumoperitoneum & 76.5  & 59.1  & 66.7  & 93.3  \\
        Pneumothorax    & 62.4  & 52.5  & 57.0  & 91.3  \\
        RFrac           & 92.6  & 83.2  & 87.7  & 98.1  \\
        VLine           & 72.2  & 69.4  & 70.8  & 91.0  \\
        NormalStudy     & 74.9  & 87.2  & 80.6  & 97.2  \\
        \bottomrule
    \end{tabular}

    \begin{tablenotes}
      \footnotesize
      \item[*]ETTube=endotracheal tube, NGTube=nasogastric tube, RFrac=rib fracture, VLine=vascular lines.
    \end{tablenotes}
    \end{threeparttable}
    \label{tab:deep_dive_table}
\end{table}

In addition to accuracy, fairness analysis was conducted to assess the model's equitable performance across gender and age groups for different adapters trained on MedImageInsight embeddings. The performance differences between female (F) and male (M) groups were minimal across all adapter models, as demonstrated by the mAUC difference shown in Table \ref{tab:fairness_gender}. Specifically, gender-related differences were small, with the largest observed mAUC disparity being 1.8\% (MLP), followed by LG (1.7\%), SVM (1.7\%), KNN (1.6\%), and RF (1.5\%). Additionally, the Wilcoxon signed-rank test revealed no significant differences between female and male patients' results, indicating robust and equitable predictions irrespective of gender. Furthermore, performance across age groups showed similarly consistent results as indicated in Table \ref{tab:fairness_age}. The standard deviation (std) of mAUC scores across the five age categories remained generally low, ranging from 0.9\% for KNN to 1.4\% for SVM, indicating minimal variability in performance across demographic groups. However, higher variability was observed within specific diagnostic classes, such as Pneumothorax (up to 6.3\% std for MLP and 4.9\% for RF), suggesting potential areas for targeted improvements. Despite these within-class discrepancies, the Kruskal-Wallis H-test shown no statistically significant differences across the five age groups. Collectively, the fairness analysis indicates that the evaluated models maintain consistent diagnostic performance across demographic subgroups, supporting equitable clinical utility.

\begin{table}[t]
  \centering
  \setlength{\extrarowheight}{1.0pt}  % Adds extra height to each row
  \small
  \begin{threeparttable}
    \caption{Gender fairness assessment of AUC performance (\%) from adapters trained by MedImageInsight embeddings. No significant gender difference was indicated by a Wilcoxon signed-rank test}
    \label{tab:fairness_gender}
    \setlength{\tabcolsep}{2.2mm}{    
        \begin{tabularx}{\textwidth}{llcccccccc}
            \toprule
            \multirow{1}{*}{Adapters} & \multirow{1}{*}{Gender} & \multirow{1}{*}{ETTube} & \multirow{1}{*}{NGTube} & \multirow{1}{*}{Pneumoperitoneum} & \multirow{1}{*}{Pneumothorax} & \multirow{1}{*}{Rrac} & \multirow{1}{*}{VLine} & \multirow{1}{*}{Normal} & \multirow{1}{*}{mAUC} \\ \midrule 
            \multirow{3}{*}{KNN} & F & 88.9 & 89.5 & 97.2 & 91.3 & 88.4 & 97.3 & 90.3 & 91.8 \\
            & M & 87.1 & 86.0 & 96.1 & 88.6 & 90.8 & 96.1 & 86.7 & 90.2 \\
            & Difference & 1.8 & 3.5 & 1.2 & 2.7 & -2.4 & 1.1 & 3.7 & 1.6 \\
            \midrule
    
            \multirow{3}{*}{LG} & F & 92.4 & 91.1 & 97.6 & 94.7 & 91.9 & 97.7 & 92.4 & 94.0 \\
            & M & 90.4 & 87.5 & 97.2 & 92.6 & 92.5 & 96.7 & 89.1 & 92.3 \\
            & Difference & 2.0 & 3.6 & 0.5 & 2.0 & -0.6 & 1.0 & 3.3 & 1.7 \\
            \midrule
    
            \multirow{3}{*}{SVM} & F & 93.0 & 91.3 & 97.9 & 94.6 & 91.1 & 98.0 & 92.6 & 94.1 \\
            & M & 90.1 & 87.7 & 97.6 & 91.9 & 92.6 & 97.0 & 89.5 & 92.4 \\
            & Difference & 2.9 & 3.6 & 0.3 & 2.7 & -1.5 & 1.0 & 3.1 & 1.7 \\
            \midrule
    
            \multirow{3}{*}{RF} & F & 91.0 & 90.4 & 97.4 & 94.4 & 90.9 & 97.4 & 91.5 & 93.3 \\
            & M & 89.3 & 86.9 & 96.8 & 92.3 & 92.0 & 96.6 & 88.3 & 91.8 \\
            & Difference & 1.7 & 3.5 & 0.6 & 2.1 & -1.1 & 0.8 & 3.1 & 1.5 \\
            \midrule
    
            \multirow{3}{*}{MLP} & F & 92.7 & 91.5 & 97.2 & 95.3 & 92.7 & 98.1 & 91.8 & 94.2 \\
            & M & 90.4 & 87.5 & 97.7 & 91.7 & 92.3 & 97.0 & 89.8 & 92.3 \\
            & Difference & 2.3 & 4.0 & -0.5 & 3.7 & 0.4 & 1.1 & 2.0 & 1.8 \\
            \bottomrule
        \end{tabularx}
    }
    \begin{tablenotes}
      \footnotesize
      \item[*] KNN = k-nearest neighbors, LG = logistic regression, SVM = support vector machine, RF = random forest, MLP = multi-layer perceptron, ETTube=endotracheal tube, NGTube=nasogastric tube, RFrac=rib fracture, VLine=vascular lines, F=female patients (3641 in total), M=male patients (5195 in total).
    \end{tablenotes}
  \end{threeparttable}
\end{table}

\begin{table}[t!]
  \centering
  \setlength{\extrarowheight}{1.0pt}  % Adds extra height to each row
  \small
  \begin{threeparttable}
    \caption{Age fairness assessment of AUC performance (\%) from adapters trained by MedImageInsight embeddings. No significant difference was observed across age groups by a Kruskal-Wallis H-Test}
    \label{tab:fairness_age}
    \setlength{\tabcolsep}{2.2mm}{    
        \begin{tabularx}{\textwidth}{llcccccccc}
            \toprule
            \multirow{1}{*}{Adapters} & \multirow{1}{*}{Age} & \multirow{1}{*}{ETTube} & \multirow{1}{*}{NGTube} & \multirow{1}{*}{Pneumoperitoneum} & \multirow{1}{*}{Pneumothorax} & \multirow{1}{*}{Rrac} & \multirow{1}{*}{VLine} & \multirow{1}{*}{Normal} & \multirow{1}{*}{mAUC} \\ \midrule 
            \multirow{5}{*}{KNN} & (0, 20] & 94.4 & 79.5 & 97.0 & 98.0 & 94.1 & 96.1 & 73.6 & 90.4 \\
            & (20, 40] & 86.7 & 89.6 & 96.4 & 89.1 & 89.5 & 95.8 & 89.6 & 91.0 \\
            & (40, 60] & 89.1 & 87.5 & 97.2 & 88.4 & 89.3 & 97.6 & 89.0 & 91.2 \\
            & (60, 80] & 86.7 & 85.7 & 96.3 & 89.8 & 90.9 & 96.6 & 88.7 & 90.6 \\
            & (80, 100] & 90.3 & 84.3 & 97.0 & 85.7 & 80.1 & 94.4 & 89.8 & 88.8 \\
            & std & 3.2 & 3.8 & 0.4 & 4.6 & 5.2 & 1.2 & 7.0 & 0.9 \\
            \midrule
    
            \multirow{5}{*}{LG} & (0, 20] & 93.3 & 84.1 & 98.4 & 98.8 & 97.0 & 99.0 & 81.2 & 93.1 \\
            & (20, 40] & 92.1 & 91.5 & 97.2 & 96.0 & 92.8 & 97.1 & 92.7 & 94.2 \\
            & (40, 60] & 91.6 & 88.9 & 97.7 & 92.3 & 90.5 & 97.6 & 90.5 & 92.7 \\
            & (60, 80] & 90.8 & 87.1 & 97.2 & 92.8 & 93.7 & 97.2 & 90.5 & 92.7 \\
            & (80, 100] & 93.3 & 87.9 & 97.3 & 86.2 & 80.9 & 96.3 & 93.2 & 90.7 \\
            & std & 1.1 & 2.7 & 0.5 & 4.7 & 6.1 & 1.0 & 4.9 & 1.3 \\
            \midrule
    
            \multirow{5}{*}{SVM} & (0, 20] & 97.9 & 84.5 & 98.4 & 95.9 & 97.1 & 99.1 & 82.0 & 93.6 \\
            & (20, 40] & 91.6 & 92.0 & 97.7 & 96.7 & 93.1 & 97.4 & 93.0 & 94.5 \\
            & (40, 60] & 91.2 & 88.8 & 97.9 & 90.8 & 89.6 & 97.9 & 90.5 & 92.4 \\
            & (60, 80] & 90.9 & 87.6 & 97.6 & 92.9 & 93.5 & 97.6 & 91.1 & 93.0 \\
            & (80, 100] & 92.7 & 88.6 & 97.7 & 86.6 & 80.9 & 96.1 & 93.4 & 90.9 \\
            & std & 2.9 & 2.7 & 0.3 & 4.1 & 6.2 & 1.1 & 4.6 & 1.4 \\
            \midrule
    
            \multirow{5}{*}{RF} & (0, 20] & 86.1 & 83.8 & 97.7 & 99.4 & 97.2 & 98.5 & 79.9 & 91.8 \\
            & (20, 40] & 90.5 & 90.9 & 97.2 & 95.5 & 92.1 & 97.0 & 91.8 & 93.6 \\
            & (40, 60] & 90.3 & 88.1 & 97.6 & 91.7 & 89.5 & 97.7 & 89.7 & 92.1 \\
            & (60, 80] & 90.3 & 86.5 & 96.8 & 92.8 & 92.6 & 97.3 & 90.0 & 92.3 \\
            & (80, 100] & 90.1 & 87.3 & 97.0 & 86.2 & 84.0 & 93.7 & 92.4 & 90.1 \\
            & std & 1.9 & 2.5 & 0.4 & 4.9 & 4.8 & 1.8 & 5.1 & 1.2 \\
            \midrule
    
            \multirow{5}{*}{MLP} & (0, 20] & 95.4 & 84.9 & 98.5 & 98.9 & 99.0 & 96.8 & 84.6 & 94.0 \\
            & (20, 40] & 90.7 & 92.9 & 97.8 & 97.7 & 91.3 & 97.8 & 93.5 & 94.5 \\
            & (40, 60] & 89.0 & 87.8 & 98.2 & 90.9 & 89.6 & 98.0 & 89.7 & 91.9 \\
            & (60, 80] & 92.2 & 87.7 & 96.2 & 92.3 & 95.4 & 97.2 & 90.4 & 93.1 \\
            & (80, 100] & 94.5 & 89.9 & 97.8 & 83.1 & 86.6 & 96.1 & 93.6 & 91.6 \\
            & std & 2.7 & 2.9 & 0.9 & 6.3 & 4.9 & 0.8 & 3.6 & 1.3 \\
            \bottomrule
        \end{tabularx}
    }
    \begin{tablenotes}
      \footnotesize
        \item[*] ETTube = endotracheal tube, NGTube = nasogastric tube, RFrac = rib fracture, VLine = vascular lines, std = standard deviation. A total of 640 patients with age in the range of (0, 20], 909 patients with age in the range of (20, 40], 2161 patients with age in the range of (40, 60], 2525 patients with age in the range of (60, 80], and 418 patients with age in the range of (80, 100] are involved in the analysis. Patients older than 100 are unidentified and are excluded from analysis.
    \end{tablenotes}
  \end{threeparttable}
\end{table}

Beyond fairness, computational efficiency was assessed via CPU-based training and inference times, as summarized in Table \ref{tab:compute_efficiency}. All models demonstrated low inference latency (<0.4 s), except SVM (2.764 s), which remained acceptable for real-time or resource-constrained clinical settings. KNN required negligible training time (0.004s), while MLP (6.374) offered a favorable balance of speed and performance. Although SVM (39.303s) and RF (155.322s) incurred higher training costs, all models remained within practical deployment limits.

These findings indicate that foundation model embeddings, particularly those derived from MedImageInsight, can be effectively utilized to train lightweight adapter classifiers, yielding high diagnostic accuracy while maintaining computational efficiency. The variability in performance across different foundation models highlights the importance of selecting appropriate embedding representations tailored to the specific radiographic classification task.

\section{Discussion}
This study evaluated the diagnostic performance and computational efficiency of lightweight adapter classifiers trained on embeddings from pre-trained foundation models for multi-class classification of radiography images, with a focus on tube placement verification and findings. Results show that adapter models effectively utilize these embeddings, achieving high mAUC values across multiple configurations. Notably, the combination of MedImageInsight embeddings with an SVM or MLP adapter achieved an mAUC of 93.1\%, outperforming all other foundation model embeddings. In contrast, Med-Flamingo embeddings yielded substantially lower mAUC values, underscoring the importance of selecting embeddings that are well-aligned with the radiographic classification task.

\begin{table}[t!]
    \centering
    \caption{Mean (standard deviation) of the training and inference time in seconds on CPU for each adapter model}
    \vspace{2mm}
    \renewcommand{\arraystretch}{1.2} % Adjust row height for better readability
    \newcolumntype{C}[1]{>{\centering\arraybackslash}p{#1}} % Define centered column type
    \begin{threeparttable}
    \begin{tabular}{l C{3.0cm} C{2.5cm} C{2.5cm} C{2.5cm} C{2.0cm}}  
        \toprule
        \textbf{Adapters} & \textbf{KNN} & \textbf{LR} & \textbf{SVM} & \textbf{RF} & \textbf{MLP} \\
        \midrule
        \textbf{Training}  & 0.004 (0.001)  & 2.913 (4.057)  & 39.303 (10.267) & 155.322 (38.407) & 6.374 (3.663) \\
        \textbf{Inference} & 0.118 (0.086)  & 0.003 (0.003)  & 2.764 (0.955)  & 0.338 (0.068)  & 0.013 (0.011)  \\
        \bottomrule
    \end{tabular}
    \begin{tablenotes}
      \footnotesize
        \item[*] KNN= K-Nearest Neighbors; LR=Logistic Regression; SVM=Support Vector Machines; RF=Random Forest; MLP=Multi-Layer Perceptron
    \end{tablenotes}
    \end{threeparttable}
\label{tab:compute_efficiency}
\end{table}

The results highlight the critical role of domain knowledge in improving classification performance. Among models trained solely on image data, Rad-DINO, which is specifically focused on radiography images, outperformed the more general DenseNet121, emphasizing the benefits of domain-focused pretraining. A similar pattern was observed among vision-language models: both BiomedCLIP and Med-Flamingo, trained on biomedical image-text pairs from scientific articles, achieved moderate performance but were outperformed by models trained on large-scale Chest-Xray imaging datasets, like CXR-Foundation. 

An especially notable finding is the exceptional performance of MedImageInsight and MedSigLIP, despite not being limited to radiography data. Unlike Rad-DINO and CXR-Foundation, which are tailored specifically to radiography and chest X-rays, MedImageInsight and MedSigLIP leverage data from a broad range of medical imaging modalities. This cross-modal training may enable the model to capture richer semantic features and shared medical imaging patterns, ultimately enhancing its representation power for radiographic tasks. These observations suggest that while task-specific domain adaptation is beneficial, diversity and scale across medical domains can provide additional gains, potentially offering more generalizable and semantically meaningful embeddings that transfer effectively across tasks. The Wilcoxon signed-rank test results further demonstrated that MedImageInsight embeddings were superior to other foundation models in this task.

Despite achieving the best overall performance, the MedImageInsight + SVM combination struggled to distinguish between ETTube, NGTube, and VLine, as evidenced by the confusion matrix in Fig. \ref{fig:confusion_matrix}. This issue is also apparent in the t-SNE visualization of MedImageInsight embeddings shown in Fig. \ref{fig:visualization}, where these three classes are not clearly separated compared to others. Such overlap suggests that while the foundation model captures general semantic features, it may not preserve the fine-grained details needed to differentiate visually similar tube placements, which may be related to the pooling operation for the embedding extraction. These findings highlight a limitation of current multi-modal foundation models, particularly in generating embeddings from spatial features through simple pooling, which might obscure subtle cues that are critical for tasks involving highly similar anatomical structures or device orientations.

The results of our investigation provide compelling evidence that foundation model embeddings, especially those derived from domain-specific models such as MedImageInsight, can be harnessed to train efficient and accurate adapter classifiers. The observed superiority of the SVM adapter across most configurations further emphasizes the potential of combining pre-trained embeddings with lightweight, traditional machine learning classifiers\cite{Paschali2025,Willemink2022}. Moreover, the computational efficiency demonstrated—where training and inference times remained within seconds for most models and around three minutes even for the more computationally demanding classifiers—supports the feasibility of integrating these methods into clinical workflows where rapid decision-making is essential.

Despite these promising findings, several limitations should be acknowledged. Although the dataset encompassed a diverse range of clinical scenarios pertinent to tube placement, future studies should aim to validate these findings in prospective cohorts and across different clinical settings. Additionally, many chest X-ray findings commonly co-occur, and a multi-label classification framework would more closely reflect clinical reality. The current single-label multi-class approach is to simplify the problem formulation and to establish a clear, reproducible benchmark task suitable for a wide range of modeling techniques. Therefore, this work is intended to serve as an initial step toward more clinically representative, multi-label classification tasks, which will be explored in future studies. Furthermore, while our study focused mainly on tube placement verification and related findings—a task of significant clinical relevance—further research is warranted to assess the applicability of this methodology to other radiographic classification challenges.

\section{Conclusion}

In summary, our investigation demonstrates that foundation model embeddings, particularly those from domain-specific sources such as MedImageInsight, can be effectively employed to train lightweight adapter models for multi‐class radiography classification with gender and age fairness. The high diagnostic accuracy and computational efficiency observed in this study underscore the potential of this approach to enhance clinical radiographic assessment, particularly in environments where rapid and reliable interpretation is critical. Future studies should further explore the integration of such models into clinical practice and assess their impact on diagnostic workflows and patient outcomes.

\section*{Acknowledgments}
This work was supported in part by the Department of Radiology at the University of Wisconsin-Madison.

\section*{Declarations}

\noindent \textbf{Ethics Approval} \quad Due to the retrospective nature of the study, a waiver of informed consent was obtained.

\noindent \textbf{Conflict of Interest} \quad NCFC, JM, MPL, ASP, NS, AE, CB are employees of Microsoft. NCFC holds diverse investments in the technology and healthcare sectors. John W. Garrett is an advisor to RadUnity. He and Joshua D. Warner are shareholders in NVIDIA. The remaining authors declare no competing interests.

%Bibliography
\bibliographystyle{sn-mathphys-num}  
\bibliography{references}

\end{document}